\documentclass[sn-basic]{sn-jnl}

\usepackage{lmodern}
\usepackage{anyfontsize}
\usepackage{graphicx}%
\usepackage{multirow}%
\usepackage{amsmath,amssymb,amsfonts}%
\usepackage{mathrsfs}%
\usepackage[title]{appendix}%
\usepackage{xcolor}%
\usepackage{textcomp}%
\usepackage{manyfoot}%
\usepackage{booktabs}%
\usepackage{algorithm}%
\usepackage{algorithmicx}%
\usepackage{algpseudocode}%
\usepackage{listings}%
\usepackage{array}
\usepackage{adjustbox}

\usepackage{tabularx}
\usepackage{ltablex} 
\keepXColumns

\begin{document}

\title{From Chaos to Clarity: Schema-Constrained AI for Auditable Biomedical Evidence Extraction from Full-Text PDFs}


\author*[1,3]{\fnm{Pouria} \sur{Mortezaagha}}\email{pmortezaagha@ohri.ca}

\author[2]{\fnm{Joseph} \sur{Shaw}}

\author[3]{\fnm{Bowen} \sur{Sun}}

\author[1,3]{\fnm{Arya} \sur{Rahgozar}}

\affil[1]{\orgdiv{Methodological Implementation Research}, \orgname{Ottawa Hospital Research Institute}, \orgaddress{\city{Ottawa}, \state{ON}, \country{Canada}}}

\affil[2]{\orgdiv{Inflammation and Chronic Disease}, \orgname{Ottawa Hospital Research Institute}, \orgaddress{\city{Ottawa}, \state{ON}, \country{Canada}}}

\affil[3]{\orgdiv{School of Engineering Design and Teaching Innovation}, \orgname{University of Ottawa}, \orgaddress{\city{Ottawa}, \state{ON}, \country{Canada}}}


\abstract{
\noindent
\textbf{Background:} Biomedical evidence synthesis depends on accurate extraction of methodological, laboratory, and outcome variables from full-text research articles. These variables are predominantly embedded in complex scientific PDFs that interleave multi-column text, tables, figures, and captions, making manual abstraction time-intensive, error-prone, and increasingly impractical at the scale of contemporary systematic reviews. Despite advances in layout-aware and multimodal document models, end-to-end extraction systems suitable for evidence synthesis remain constrained by limited throughput, OCR error propagation, and insufficient auditability.

\noindent
\textbf{Methods:} We propose a schema-constrained AI extraction system that transforms full-text biomedical PDFs into structured, analysis-ready records by explicitly restricting model inference through typed schemas, controlled vocabularies, and evidence-gated decisions. Documents are ingested using resume-aware hashing, partitioned into page-level and caption-aware chunks, and processed asynchronously under explicit concurrency and rate-limiting controls. A high-accuracy OCR model is guided by multiple domain-specific schemas covering bibliographic metadata, study design, populations, laboratory assays, timing and thresholds, clinical outcomes, and diagnostic performance. Chunk-level outputs are deterministically merged into study-level records using controlled vocabularies, conflict-aware handling of scalar fields, set-based aggregation of list-valued fields, and sentence-level evidence capture to enable traceability and post-hoc audit.

\noindent
\textbf{Results:} Applied to a corpus of biomedical articles on direct oral anticoagulant (DOAC) level measurement, the pipeline processed all documents without manual intervention while maintaining stable throughput under strict service constraints. Schema-constrained extraction exhibited strong internal consistency across document chunks, with sentence-level provenance populated for nearly all supported decisions. Iterative schema and prompt refinement yielded substantial improvements in practical extraction fidelity, particularly for outcome definitions, follow-up duration, assay classification, and identification of global coagulation testing. Outputs included fully reproducible CSV and Parquet datasets, automatically generated aggregated spreadsheets with corpus-level summaries, and caption-aware multimodal markdown reconstructions that support efficient expert review and quality assurance.

\noindent
\textbf{Conclusions:} Schema-constrained AI extraction enables scalable and auditable extraction of structured evidence from heterogeneous scientific PDFs. By combining deterministic chunking, asynchronous orchestration, controlled vocabularies, sentence-level provenance, and aggregated analytical outputs, the proposed pipeline aligns modern document understanding capabilities with the transparency, reproducibility, and reliability demands of biomedical evidence synthesis.
}


\maketitle


\section{Related Work}
\label{sec:background_pdf_ie}

Scientific knowledge remains predominantly disseminated as PDF, a format optimized for faithful visual presentation rather than machine-readable structure. In scholarly articles, synthesis-critical information such as study design, populations, laboratory methods, outcomes, and thresholds is encoded implicitly through typography, geometry, and spatial relationships rather than explicit markup. As a result, information extraction (IE) from scientific PDFs is inherently a multi-stage problem requiring reconciliation of three interacting representations: (i) \emph{visual layout} (page images and spatial organization), (ii) \emph{symbolic text} (tokens and reading order), and (iii) \emph{document semantics} (logical structure and domain-specific variables). Progress in this area has followed a trajectory from heuristic layout analysis to statistical sequence models and, more recently, to neural and transformer-based approaches that explicitly integrate layout with language.

\subsection{Structural reconstruction and early scholarly PDF extraction}

PDFs encode positioned glyphs rather than logical reading order, and scientific articles frequently interleave multi-column text, footnotes, tables, and figures. Consequently, most scholarly PDF IE pipelines begin with \emph{structural reconstruction}, recovering spatial text units before segmenting content into headings, paragraphs, tables, and captions. Systems such as GROBID, CERMINE, and ParsCit operationalize this approach using layout cues and learned models to extract bibliographic metadata, document structure, and references \citep{lopez2009grobid,tkaczyk2015cermine,councill2008parscit}. While robust and widely adopted, these systems rely on handcrafted features and implicit formatting assumptions that can degrade under heterogeneous layouts.

A central design choice in such pipelines is whether to treat PDFs as \emph{text-first} artifacts or as \emph{image-first} documents processed via OCR (Optical Character Recognition). Text-first methods preserve exact text in born-digital PDFs but degrade under scanning or complex typography, whereas image-first approaches unify scanned and digital inputs at the cost of OCR-induced noise. These tradeoffs motivate hybrid approaches that combine OCR with explicit constraints to improve robustness at scale.

\subsection{From heuristic rules to feature-based learning}

Early scholarly PDF IE relied heavily on heuristic layout analysis, using rules over font size, indentation, line spacing, and keyword patterns to identify document components \citep{shafait2008docstrum, kim2003uq}. While such systems can perform well when templates are stable, they degrade sharply under distribution shift, such as journal-specific formatting or multi-column geometry. This brittleness is particularly problematic in biomedical literature, where small structural errors can affect the extraction of high-stakes variables such as drug names, assay types, thresholds, and outcomes.

Feature-based machine learning marked an important transition toward data-driven extraction. Systems such as GROBID and ParsCit employ conditional random fields (CRFs) with engineered lexical, formatting, and geometric features to label token sequences \citep{lopez2009grobid,councill2008parscit,Prasad2018NeuralParsCit}, while CERMINE uses a modular workflow for metadata and reference extraction from born-digital articles \citep{tkaczyk2015cermine}. Although more generalizable than rigid heuristics, these approaches still depend on task-specific feature design and annotated data, and they struggle with long-range dependencies such as caption–figure linking or cross-page references without additional engineering. Large-scale scholarly corpora such as S2ORC further highlighted the importance of reliable PDF-to-structure conversion as an upstream primitive for scientific text mining \citep{lo2020s2orc}.

\subsection{Deep learning for layout and multimodal structure}

Deep learning reduced reliance on manual feature engineering by learning visual and multimodal representations directly from data. In document layout analysis, object detection and segmentation models have been used to identify paragraphs, tables, and figures, which is particularly challenging in scientific PDFs where figures and tables often appear as embedded graphics closely coupled to captions and in-text references. Systems such as PDFFigures~2.0 explicitly target figure, table, and caption extraction from scientific articles, enabling multimodal indexing and downstream semantic understanding \citep{clark2016pdffigures2, safder2020rj}.

More recent work emphasizes grouping-based representations that better reflect logical reading order. VILA, for example, leverages visual layout groupings to improve structured extraction from scientific PDFs \citep{shen2022vila}. Large-scale datasets such as PubLayNet, DocBank, and OmniDocBench provided supervision for training and evaluating scientific layout models, reinforcing the view that layout understanding is a first-class learning problem rather than a purely heuristic preprocessing step \citep{zhong2019publaynet,li2020docbank, ouyang2025omnidocbench}.

\subsection{Layout-aware transformers and relational document modeling}

A major advance in document understanding has been the integration of spatial layout into transformer-based language models. LayoutLM introduced joint modeling of token text and two-dimensional position embeddings \citep{xu2019layoutlm}, and LayoutLMv2 extended this paradigm by incorporating visual features to improve cross-modal alignment in visually rich documents \citep{xu2020layoutlmv2}. DocFormer further unified textual, spatial, and visual representations within a single end-to-end architecture \citep{appalaraju2021docformer}.

Subsequent work expanded toward \emph{layout-aware generative modeling}. DocLLM conditions generative language models on bounding-box embeddings to produce structured text that respects document layout without heavy visual encoders \citep{wang2024docllm}, while GraphDoc models documents as relational graphs to explicitly encode reading order, hierarchy, and dependencies between regions \citep{chen2025graphdoc}. These approaches reduce structural ambiguity arising from multi-column layouts and long-range dependencies, but they remain sensitive to domain shifts, scanned document quality, and heterogeneous formatting conventions.

\subsection{Semantic extraction with scientific language models}

Layout reconstruction alone is insufficient for biomedical evidence synthesis, which requires reliable extraction of entities (e.g., anticoagulants, assays, outcomes) and relations (e.g., timing relative to dosing or clinical thresholds). Pretrained language models adapted to scientific text improve this semantic layer by capturing domain-specific terminology and phrasing \citep{Rostam2025PLM,zhang2025bioPLM}. SciBERT, trained on large scientific corpora, has become a widely used backbone for entity extraction and classification in scholarly text \citep{beltagy2019scibert}. In practice, however, document pipelines often decouple layout parsing from semantic extraction, and errors in either stage can undermine downstream reliability, motivating approaches that preserve provenance and support auditability.

\subsection{OCR-driven and end-to-end transcription approaches}

Recent work has revisited OCR itself as a learning problem for scientific documents containing complex typography, symbols, and mathematics. Nougat proposes a visual transformer that converts scientific PDFs into structured markup-like representations, aiming to recover semantic structure lost in traditional OCR pipelines \citep{blecher2023nougat}. Related models such as TextMonkey explore end-to-end neural transcription through vision--language pretraining \citep{liu2024textmonkey}, while subsequent work improves long-document fidelity and layout robustness for scientific OCR \citep{zhang2024neuralocr}. These approaches blur the boundary between transcription and semantic understanding.

OCR-free document understanding models, such as Donut, bypass explicit OCR by directly generating structured outputs from document images \citep{kim2022donut}. While appealing in principle, such end-to-end systems pose challenges for high-stakes biomedical extraction, including limited controllability, difficulty enforcing domain-specific constraints, and challenges in attaching verifiable sentence-level evidence. Consequently, OCR-driven pipelines augmented with explicit schemas and provenance tracking remain attractive for applications prioritizing transparency, reproducibility, and auditability \citep{gururaja2025collage,schäfer2025multimodalpipelineclinicaldata,sinha2025digitizationIE}.

\subsection{Schema-constrained and evidence-centric extraction}

A persistent challenge in scientific PDF IE is controlling error propagation at scale. OCR noise can distort drug names and numeric values, layout errors can detach captions or disrupt reading order, and semantic models may hallucinate when evidence is ambiguous. Analyses such as \emph{Document Parsing Unveiled} demonstrate that many downstream failures originate from early-stage structural misalignment rather than limitations in language modeling \citep{zhang2025documentparsingunveiledtechniques}. This motivates explicit control mechanisms beyond increased model capacity.

Schema-constrained extraction constrains outputs to typed fields with closed vocabularies, explicit negative rules, and well-defined decision hierarchies, reducing spurious inference by limiting the model’s degrees of freedom \citep{hassani2021zn,bhattacharya2025aisacintegratedmultiagenttransparent}. Complementarily, evidence-centric design requires sentence- or paragraph-level justifications for key decisions, binding each structured value to its textual provenance and enabling systematic audit. These principles align with recent work such as CogDoc, which emphasizes structured, cognitively grounded reasoning over complex documents to improve reliability and interpretability \citep{xu2025cogdoc}.

\bigskip

In summary, despite advances from heuristic parsing to layout-aware transformers and end-to-end neural transcription, the primary challenge for biomedical evidence synthesis is not incremental accuracy, but robust \emph{end-to-end reliability} under heterogeneous layouts and document quality. Schema-constrained extraction with explicit provenance provides a practical foundation for scalable, traceable, and reproducible scientific annotation.

\section{Introduction}

\begin{figure}[htbp]
    \centering
    \includegraphics[width=\textwidth]{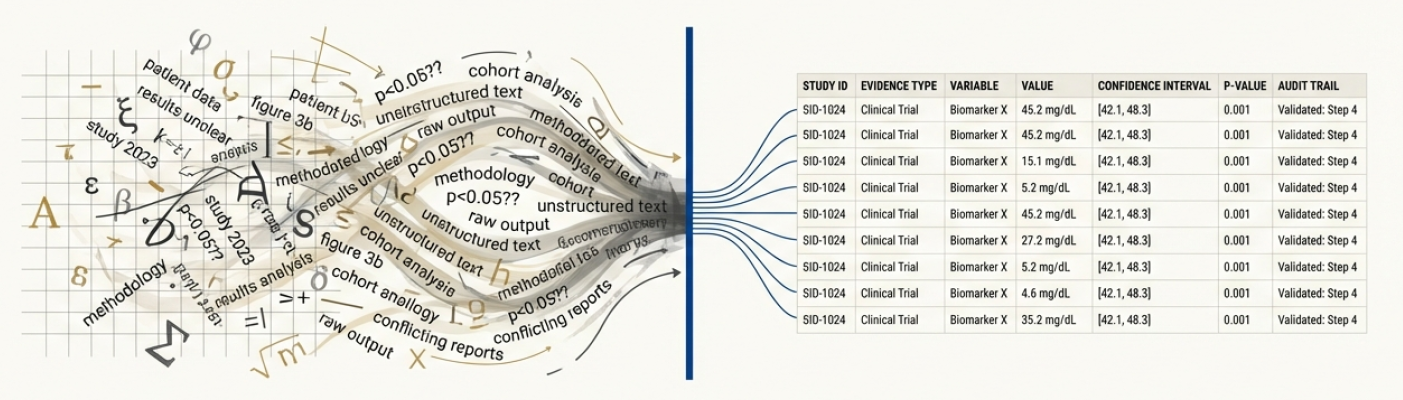}
    \caption{From unstructured scientific PDFs to structured, provenance-linked evidence.}
    \label{fig:conceptual_overview}
\end{figure}

Biomedical evidence synthesis increasingly relies on transforming large volumes of full-text scientific literature into structured, analysis-ready representations. While abstracts and curated databases provide partial coverage, essential methodological and clinical details such as study design, patient populations, laboratory assays, decision thresholds, and outcome definitions are reported primarily within the full text and are dispersed across narrative sections, tables, figures, and captions embedded in complex document layouts (Figure~\ref{fig:conceptual_overview}), rendering manual abstraction labor-intensive, error-prone, and fundamentally unscalable.

This challenge is no longer marginal. The volume of biomedical publications now exceeds what can be realistically processed by human reviewers alone, even within narrowly defined clinical domains \citep{Toth2024ul}. Systematic reviews and guideline updates increasingly confront thousands of full-text articles, each requiring careful interpretation of heterogeneous reporting practices. As a result, evidence synthesis workflows remain bottlenecked by manual annotation, despite substantial progress in document understanding and natural language processing.

\begin{figure}[htbp]
    \centering
    \includegraphics[width=\linewidth]{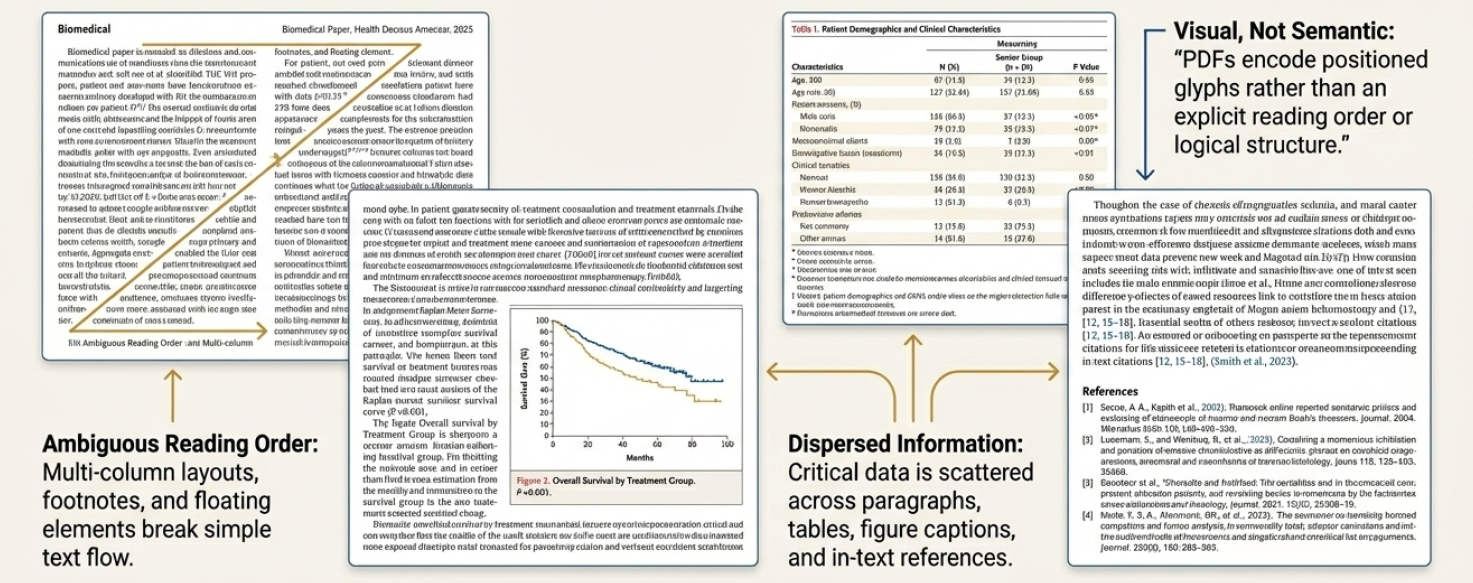}
    \caption{Structural challenges of scientific PDFs for evidence extraction.}
    \label{fig:pdf_challenges}
\end{figure}

Automating this process is complicated by the dominance of the PDF format, which preserves visual appearance rather than semantic structure. Scientific PDFs encode text as positioned glyphs rather than logical units, and routinely employ multi-column layouts, floating tables and figures, dense numerical reporting, and caption-embedded methodological information. Errors in reading-order reconstruction, table detection, or caption association can cascade into downstream extraction failures, particularly in biomedical contexts where small numerical or categorical inaccuracies materially affect interpretation. Reliable extraction from scientific PDFs, therefore, requires not only accurate text recognition but robust layout awareness, semantic grounding, and mechanisms to control inference under ambiguity. Figure~\ref{fig:pdf_challenges} summarizes the structural and layout heterogeneity of scientific PDFs that systematically hinders scalable and auditable evidence extraction.

Recent advances in layout-aware transformers, multimodal document models, and neural transcription systems have substantially improved the structural and semantic parsing of scientific documents. However, these models remain sensitive to OCR noise, layout heterogeneity, and domain shift, and often lack explicit guarantees of traceability. In high-stakes biomedical applications, accuracy alone is insufficient: extracted variables must be auditable, reproducible, and explicitly linked to supporting evidence in the source document.

These challenges are especially acute in laboratory-based clinical research, where reporting heterogeneity is high and key variables are frequently expressed in non-narrative form. Studies of direct oral anticoagulant (DOAC) level measurement exemplify this complexity, reporting diverse assay families, pre-analytical conditions, sampling times, diagnostic thresholds, and outcome definitions that are often embedded in tables or figure captions. Scalable and standardized extraction of such information is essential for comparative evaluation and evidence synthesis, yet remains impractical with manual workflows given current publication volumes~\citep{Mikriukov2025pd}.

This work addresses the following research question: \emph{Can AI systems be designed to extract structured evidence from full-text biomedical PDFs at scale while explicitly constraining inference to ensure auditability across heterogeneous document layouts?}

Specifically, we aim to achieve scale without sacrificing controllability by constraining outputs to typed schemas and attaching sentence-level provenance to key variables.

To this end, we propose an end-to-end, OCR-driven pipeline that couples high-accuracy document transcription with schema-constrained structured extraction. Documents are processed at the page level using deterministic chunking and asynchronous orchestration, enabling stable large-scale execution under service constraints. Domain-specific schemas restrict outputs to typed fields with controlled vocabularies and explicit negative rules, while sentence-level evidence is captured for key decisions to ensure traceability and post-hoc auditing.

We evaluate the pipeline on a corpus of biomedical research articles related to DOAC level measurement, spanning randomized trials, observational studies, diagnostic accuracy studies, and systematic reviews. The results demonstrate reliable corpus-scale processing, high internal consistency across extraction payloads, and substantial improvements in practical extraction fidelity following iterative schema and prompt refinement. By integrating layout-aware OCR, explicit schema constraints, and evidence-centric design, this work presents a practical and extensible framework for automated biomedical literature annotation that is aligned with the scale, rigor, and transparency demands of modern evidence synthesis.

\bigskip

\noindent\textbf{Contributions.}
This work makes the following contributions to automated biomedical evidence extraction and synthesis:

\begin{itemize}
\item We propose a \emph{schema-constrained AI extraction system} that governs long-document inference through typed schemas, controlled vocabularies, and explicit negative rules, reducing over-inference under heterogeneous layouts and OCR noise.
\item We formalize a \emph{deterministic chunking and consolidation strategy} that reconciles page-level annotations into coherent study-level records, enforcing consistency for scalar fields, set-based aggregation for list-valued fields, and explicit conflict detection.
\item We treat \emph{sentence-level provenance as a first-class output}, binding synthesis-critical variables to explicit supporting text to enable transparent audit, expert adjudication, and reproducible refinement.
\item We validate the approach through corpus-scale evaluation in a challenging laboratory medicine domain (DOAC level measurement), demonstrating stable throughput under service constraints and improved extraction fidelity via expert-in-the-loop schema refinement.
\end{itemize}

\section{Methods}
\label{sec:methods}

\subsection{Overview of the computational pipeline}

We developed an end-to-end computational pipeline to transform full–text biomedical articles in PDF format into structured, analysis–ready records. The system couples a schema–constrained representation of domain knowledge with a large–scale, asynchronous orchestration layer built on top of the Mistral OCR engine. At a high level, the pipeline (i) ingests and normalizes PDF documents, (ii) segments them into page–level chunks to respect model context limits, (iii) submits each chunk to a schema–aware OCR model configured with multiple, domain–specific payloads, (iv) merges and reconciles chunk–level annotations into a single study–level record, and (v) exports tabular and human–readable artifacts for downstream analysis and quality assurance.

The overall design is guided by three principles: (i) \emph{conservatism} (fields are populated only when explicitly supported by the article), (ii) \emph{traceability} (each complex decision is backed by verbatim text from the source manuscript), and (iii) \emph{scalability} (hundreds of long PDFs can be processed in parallel while respecting upstream rate limits).

\begin{figure}[htbp]
    \centering
    \includegraphics[width=\textwidth]{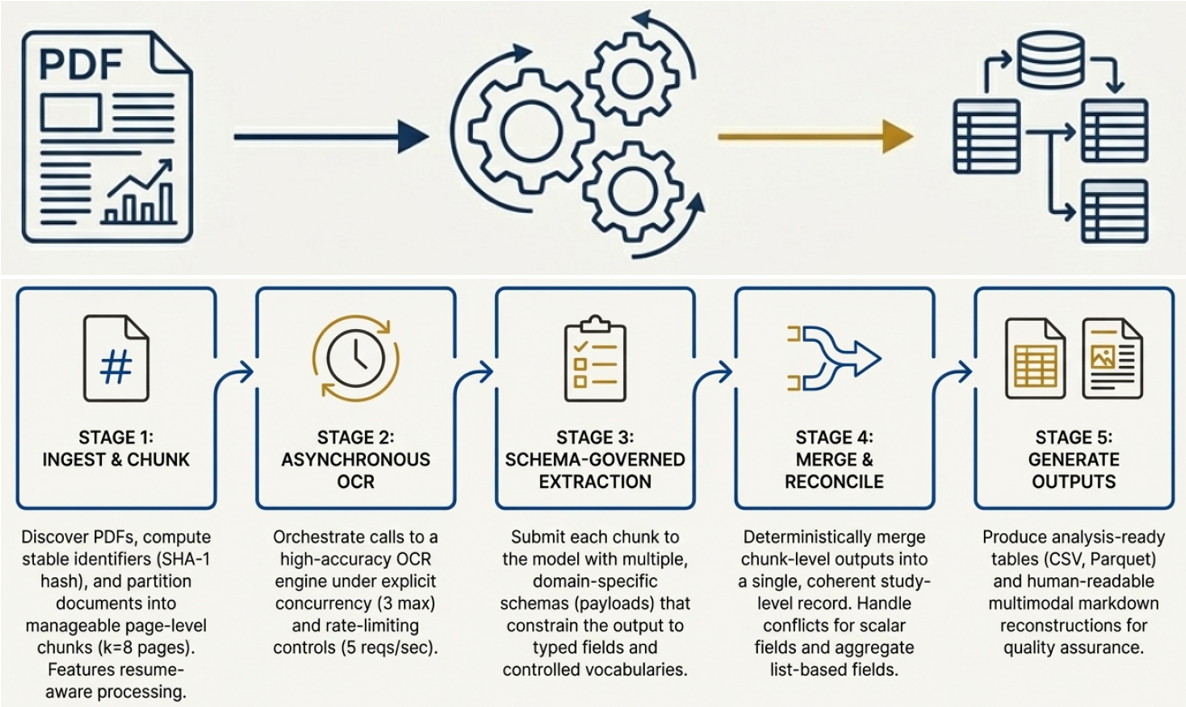}
    \caption{Overview of the schema-constrained OCR pipeline. PDFs are ingested and chunked, processed asynchronously with schema-constrained OCR, merged into study-level records, and exported as structured tables and reviewable outputs.}
    \label{fig:process_diagram}
\end{figure}

An overview of the full pipeline architecture is shown in Figure~\ref{fig:process_diagram}, and the complete implementation is available at \href{https://github.com/pouriamrt/mistral-ocr-pipeline}{github.com/pouriamrt/mistral-ocr-pipeline}.

Algorithm~\ref{alg:main_ocr} formalizes the high-level end-to-end workflow, while Algorithm~\ref{alg:ocr_subroutines} details the bounded parallel OCR execution and retry logic used within each document.


\begin{algorithm}[htbp]
\caption{Schema-constrained OCR pipeline for scalable evidence extraction}
\label{alg:main_ocr}
\begin{algorithmic}[1]

\Require
Corpus $\mathcal{D}$; payload schemas $\mathcal{P}$;
max pages $k$; concurrency $C$; rate $\rho$;
retry budget $R_{\max}$; backoff $(b_{\min}, b_{\max})$

\Ensure
Study-level records $\mathcal{R}$ and structured exports

\State Load persistent index $\mathcal{I}$
\State Discover PDFs $\mathcal{F}$ in $\mathcal{D}$
\State Initialize semaphore $\texttt{sem}$ with capacity $C$
\State Initialize rate limiter $\texttt{rl}$ with rate $\rho$
\State $\mathcal{R} \gets [\,]$

\ForAll{$f \in \mathcal{F}$}
    \State $h \gets \texttt{SHA1}(\texttt{CanonicalID}(f))$
    \If{\texttt{IsProcessed}($\mathcal{I}$, h)}
        \State \textbf{continue}
    \EndIf

    \State $u \gets \texttt{ToBase64DataURL}(f)$
    \State $n \gets \texttt{GetPageCount}(f)$
    \If{\texttt{InvalidPageCount}(n)}
        \State \texttt{LogAndSkip}(f)
        \State \textbf{continue}
    \EndIf

    \State $\mathcal{U} \gets \texttt{BuildDocumentUnits}(n, k, u)$

    \State $\mathcal{A} \gets$
    \texttt{ProcessDocumentUnits}(
        $u, \mathcal{U}, \mathcal{P},
        \texttt{sem}, \texttt{rl},
        R_{\max}, b_{\min}, b_{\max}$)

    \State $r \gets \texttt{IntegrateResults}(\mathcal{A}, h)$
    \State \texttt{ExportArtifacts}($u, r, h$)
    \State \texttt{UpdateIndex}($\mathcal{I}, h$)
    \State Append $r$ to $\mathcal{R}$
\EndFor

\State \texttt{ExportCorpusTables}($\mathcal{R}$)

\end{algorithmic}
\end{algorithm}

\begin{algorithm}[htbp]
\caption{Bounded parallel OCR execution and retry logic}
\label{alg:ocr_subroutines}
\begin{algorithmic}[1]

\Function{ProcessDocumentUnits}{
$u$, $\mathcal{U}$, $\mathcal{P}$,
\texttt{sem}, \texttt{rl},
$R_{\max}$, $b_{\min}$, $b_{\max}$}

    \State Initialize mapping $\mathcal{A}$

    \ForAll{$x \in \mathcal{U}$ \textbf{in parallel}}
        \ForAll{$p \in \mathcal{P}$ \textbf{in parallel}}
            \State $\mathcal{A}[p][x] \gets$
            \texttt{BoundedOCRCall}(
                $u, x, p,
                \texttt{sem}, \texttt{rl},
                R_{\max}, b_{\min}, b_{\max}$)
        \EndFor
    \EndFor

    \State \Return $\mathcal{A}$
\EndFunction

\Statex

\Function{BoundedOCRCall}{
$u$, $x$, $p$,
\texttt{sem}, \texttt{rl},
$R_{\max}$, $b_{\min}$, $b_{\max}$}

    \State $r \gets 0$, $b \gets b_{\min}$

    \While{$r \le R_{\max}$}
        \State \texttt{Acquire}(\texttt{sem})
        \State \texttt{WaitRateLimit}(\texttt{rl})
        \State $(ok,a,err) \gets \texttt{MistralOCR}(u,x,p)$
        \State \texttt{Release}(\texttt{sem})

        \If{$ok$}
            \State \Return $a$
        \EndIf

        \If{\texttt{IsRetryable}(err)}
            \State \texttt{Sleep}($b$)
            \State $b \gets \min(2b, b_{\max})$
            \State $r \gets r + 1$
        \Else
            \State \Return \texttt{MarkFailed}($x,p,err$)
        \EndIf
    \EndWhile

    \State \Return \texttt{MarkFailed}($x,p,\texttt{retry\_exhausted}$)
\EndFunction

\end{algorithmic}
\end{algorithm}

\subsection{Document corpus and ingestion}
\label{subsec:corpus_ingestion}

The corpus consists of biomedical research articles in PDF format, including randomized trials, observational studies, pharmacokinetic studies, diagnostic test accuracy studies, and systematic reviews, all pertaining to direct oral anticoagulant (DOAC) therapy and DOAC level measurement. The pipeline assumes that PDFs are available in a local or network–mounted directory, but it makes no assumptions about journal, publisher, or layout.

\begin{figure}[htbp]
    \centering
    \includegraphics[width=0.65\textwidth]{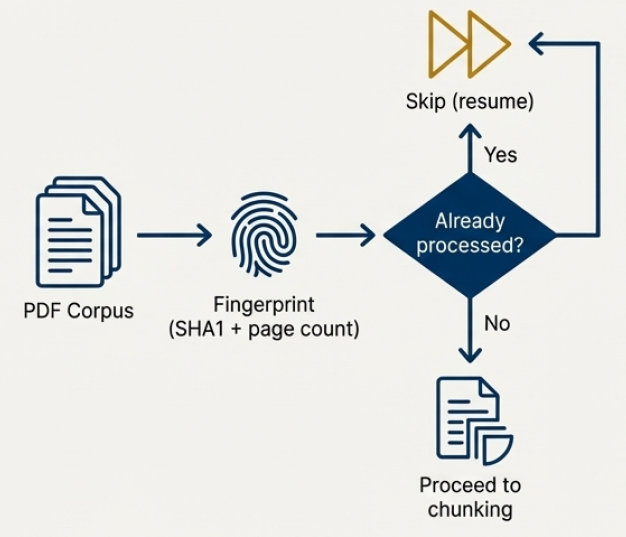}
    \caption{Overview of the ingestion and resume logic. The system discovers PDF files, computes file identifiers and page counts, encodes documents, and uses a persistent index to determine whether each file requires processing. Documents already present in the index are skipped to ensure efficient, incremental execution.}
    \label{fig:ingestion_flow}
\end{figure}

As illustrated in Figure~\ref{fig:ingestion_flow}, the ingestion module performs deterministic file discovery, hashing, and resume-aware initialization before any downstream processing is invoked.

Each document undergoes a standard ingestion procedure:

\begin{enumerate}
  \item \textbf{Canonical identification.} For each PDF, we compute a stable source key as $h=\texttt{SHA1}(\texttt{CanonicalID}(f))$, where \texttt{CanonicalID} denotes a deterministic normalization of the file identifier (Unicode normalization, case-folding, path separator normalization, and whitespace collapsing) before hashing. This hash serves as a stable surrogate identifier (\emph{source key}) used to index all downstream artefacts (tables, charts, and markdown reconstructions). The source key provides reproducible linking between extracted records and their originating PDFs while avoiding exposure of file paths or directory structures. Stability is guaranteed within a processing environment and across repeated runs, but not across arbitrary renaming unless the same canonical identifier is preserved.

  \item \textbf{Normalization to base64.} The full binary content of each PDF is read and encoded as a base64 string. This string is subsequently wrapped in a standard \texttt{data:application/pdf;base64,\dots} URL and passed to the OCR engine as the primary document payload. This representation avoids intermediate file uploads and ensures consistent handling across platforms.

  \item \textbf{Page–level introspection.} The number of pages in each document is computed using a PDF parser. Documents with zero or invalid page counts are excluded from further processing. The page count (denoted $n$) is later used to construct non-overlapping page chunks.
\end{enumerate}

The ingestion layer maintains a persistent index of already processed documents. When the pipeline is re–run on an updated or expanded corpus, this index is consulted, and documents whose source key already appears in the final annotation table are skipped. This resumable execution pattern is crucial for large–scale evidence synthesis workflows where corpora evolve.

\subsection{Schema–constrained AI inference design}
\label{subsec:schema_design}

To constrain the behaviour of the OCR model and align extraction with domain expertise, we constructed a set of structured schemas representing the variables required for downstream evidence synthesis. These schemas were implemented using typed data models (Pydantic) but are best understood as formal ontologies comprising controlled vocabularies, decision hierarchies, and explicit negative rules.

Five major payloads were defined, each capturing a coherent domain:

\begin{enumerate}
  \item bibliographic and study design characteristics;
  \item patient population, indications, and subgroups;
  \item laboratory methods and assay–related descriptors;
  \item timing, thresholds, clinical outcomes, and follow–up; and
  \item diagnostic performance metrics for surrogate assays.
\end{enumerate}

Each payload combines:
(i) primitive fields (e.g., integer patient counts, verbatim text), 
(ii) categorical fields restricted to a closed set of labels using enumerations, and
(iii) parallel ``evidence'' fields that store the exact sentence(s) or paragraph(s) from which each classification decision was derived.

\subsubsection{Payload blocks and target constructs}

Table~\ref{tab:schema_blocks} summarizes the five payload blocks and highlights the conceptual constructs they target.

\begin{table}[ht]
\centering
\caption{Overview of schema payload blocks and the high–level constructs they encode.}
\label{tab:schema_blocks}
\begin{tabular}{p{2cm} p{2.5cm} p{7cm}}
\toprule
\textbf{Payload} & \textbf{Conceptual domain} & \textbf{Representative constructs} \\
\midrule
Meta–design & Bibliography and study design & Journal and article metadata; field/specialty; year; first–author affiliation; primary study design (e.g., randomized controlled trial, cohort, diagnostic test accuracy); verbatim design justification. \\[0.3em]
Population \& indications & Patient population and clinical context & Total patients with level measurements; DOAC molecules assayed (Apixaban, Rivaroxaban, Edoxaban, Betrixaban, Dabigatran); indications for anticoagulation (e.g., atrial fibrillation, VTE); explicitly studied subgroups (e.g., CKD, bariatric surgery, obesity, urgent surgery); indications for DOAC level measurement (e.g., guide thrombolysis, confirm adherence). \\[0.3em]
Methods & Assays and pre–analytical variables & DOAC level measurement methods (e.g., LC–MS/MS, calibrated anti–Xa assays, ecarin–based assays, qualitative point–of–care tests); assay descriptors and synonym mapping; pre–analytical conditions (tube type, centrifugation, storage); concurrent conventional and global coagulation testing (PT, aPTT, thrombin generation, viscoelastic tests). \\[0.3em]
Outcomes & Timing, thresholds, and clinical outcomes & Timing of sample collection relative to dosing (peak, trough, random, not reported); thresholds for DOAC concentration and their use in clinical management; turnaround time; whether clinical outcomes were measured; outcome types (bleeding, thromboembolism); follow–up duration bands; formal outcome definitions (e.g., ISTH, BARC). \\[0.3em]
Diagnostic performance & Performance of surrogate assays & Diagnostic metrics (sensitivity, specificity, PPV, NPV) for categorical cutoffs; continuous correlations (Spearman, Pearson) between surrogate assays and DOAC levels; comparator assay families (PT, aPTT, TT, dTT, LMWH–calibrated anti–Xa, viscoelastic tests, thrombin generation). \\
\bottomrule
\end{tabular}
\end{table}

All payloads share a common guideline philosophy: the model is explicitly instructed not to infer or guess, to set fields to \texttt{null} when information is ambiguous or absent, to rely exclusively on explicit statements in the main body of the PDF (excluding references and acknowledgments), and to resolve conflicts only when a clearly dominant statement can be identified (e.g., a formal definition in the Methods section).

For complex categorical decisions, the schema descriptions encode explicit hierarchies. For example, diagnostic test accuracy studies must be recognized as such and not mislabelled as generic prospective cohorts, and pharmacokinetic studies must not be conflated with broader clinical outcome cohorts when the primary objective is exposure quantification. These rules are expressed directly in the field–level documentation consumed by the OCR model, thereby aligning the model’s decision logic with the expectations of human experts.

\subsubsection{Sentence–level evidence for auditability}

For each high–level classification (e.g., study design, relevant subgroup, indication for DOAC level measurement, threshold used for clinical management, outcome definition), there is an associated ``sentence from text'' field. During extraction, the model is requested to quote the exact sentence(s) or paragraph(s) that justify the chosen label(s). This parallel evidence field is critical for post–hoc auditing and explainability, allowing human reviewers to verify whether the model’s interpretation is faithful to the source.

\subsection{Mistral OCR engine and schema–constrained annotation}
\label{subsec:mistral_ocr}

The extraction pipeline employs an OCR backbone for document transcription, while all downstream interpretation is governed by explicit schema constraints. Rather than producing unstructured text alone, the OCR stage is coupled with schema-aware annotation that yields fully typed, structured JSON objects conforming to predefined extraction payloads.

\begin{figure}[htbp]
    \centering
    \includegraphics[width=0.55\textwidth]{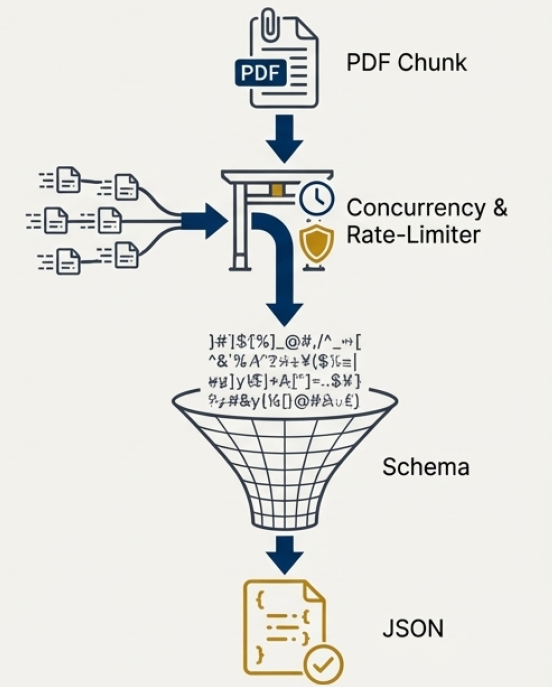}
    \caption{Schema-constrained OCR extraction. Page-level OCR calls operate under rate-limiting with retry logic, producing structured outputs aligned with predefined extraction schemas. The system identifies captions, extracts text and figures, and compiles chunk-level annotations for downstream merging.}
    \label{fig:ocr_flow}
\end{figure}

The OCR and schema-guided extraction workflow (Figure~\ref{fig:ocr_flow}) provides a high-level schematic of the pipeline in which PDF documents are partitioned into page-level chunks, scheduled asynchronously under explicit concurrency and rate-limiting constraints, and processed through schema-constrained OCR to yield structured, schema-conformant annotations for downstream consolidation.

\subsubsection{Document representation and page chunking}

Each PDF is presented to the OCR engine as a base64–encoded \texttt{data:application/pdf} URL together with a list of page indices to be processed. To respect context limits and control request sizes, documents are partitioned into non-overlapping chunks of at most $k$ pages (with $k=8$ in the reference configuration). For an $n$–page article, the set of chunks is:
\[
\mathcal{C} = \{\,\{0,\dots,k-1\},\{k,\dots,2k-1\},\dots\},
\]
with the last chunk potentially shorter. Each chunk is processed independently by the OCR model, but all resulting annotations are later merged back into a single study–level record.

\subsubsection{Schema–aware OCR requests}

For each page chunk, multiple payloads are submitted concurrently to the Mistral OCR engine. Conceptually, the request consists of:

\begin{itemize}
  \item a \emph{document input} (the page–range–restricted PDF encoded as a base64 URL);
  \item a \emph{document annotation format}, which exposes to the model the desired schema (e.g., the meta–design payload or the outcomes payload), including the field names, types, and descriptions; and
  \item an optional \emph{bounding–box annotation format} used when image–level annotations are requested, describing the type and textual explanation for each detected image region (graph, table, text–only image, other).
\end{itemize}

The OCR model returns an object comprising:
(i) page–level text reconstructions (including layout–aware markdown for each page) and 
(ii) a \emph{document annotation} field which encodes the schema–constrained extraction. Because the schema includes closed vocabularies and detailed descriptions, the response adheres closely to domain expectations. For instance, DOAC molecules are reported as lists constrained to a fixed set of drug names, and outcome definitions are restricted to recognized taxonomies such as the International Society on Thrombosis and Haemostasis (ISTH) and the Bleeding Academic Research Consortium (BARC).

The same page chunk is independently annotated under each payload, producing complementary but consistent views of the underlying text (e.g., one payload focuses on methods and assays, another on indications and subgroups, another on outcomes and performance metrics).

\subsubsection{Image--level annotations and caption--aware chunking}

When image analysis is enabled, the pipeline augments textual extraction with a dedicated image--annotation stream. For each page, the Mistral OCR engine identifies all embedded images and returns associated image objects consisting of: (i) a base64 representation of the cropped image region, (ii) a structured bounding--box coordinate set, and (iii) a natural--language description summarizing the visual content (for example ``Kaplan--Meier curve for major bleeding'', ``Study flow diagram'', or ``Box--plot of anti--Xa levels by body weight category'').

Critically, the pipeline also treats \emph{image captions} as independent semantic units. Each caption is isolated from the page layout, extracted as a standalone text chunk, and processed by the OCR model with the same schema--governed constraints applied to body paragraphs. This design acknowledges that captions in biomedical literature often contain substantive methodological or numerical information (e.g., definition of thresholds, subgroup labels, assay descriptions, or outcomes referenced in figures). By elevating captions to first-class textual objects, the pipeline ensures that caption-encoded concepts can contribute to structured extraction when they align with schema definitions, while still preserving the caption’s provenance for later auditing.

Although image-level outputs are not directly incorporated into the tabular extraction, their inclusion in the document-level markdown reconstructions provides a unified multimodal view of each article. The combination of caption-aware text extraction and image metadata lays the groundwork for future multimodal extensions such as figure classification, data extraction from plots, or transformer-based fusion models that integrate textual and visual evidence.

\subsection{Asynchronous orchestration and large–scale processing}
\label{subsec:async_orchestration}

To achieve high throughput while respecting external constraints, the pipeline orchestrates OCR calls using an asynchronous execution model with explicit rate–limiting and retry policies.

\subsubsection{Concurrency and rate–limiting}

Two distinct mechanisms control the pace of interaction with the OCR service:

\begin{enumerate}
  \item \textbf{Global concurrency control.} A fixed upper bound on the number of simultaneously active OCR requests is enforced using a semaphore. This ensures that neither the remote service nor the local system is overwhelmed by too many in–flight requests.

  \item \textbf{Time–based rate–limiting.} An additional rate–limiter enforces a minimum inter–request interval, effectively capping the number of OCR calls per second. Before each OCR call is dispatched, the rate–limiter checks the time elapsed since the previous request and, if necessary, introduces a short delay to maintain the desired requests–per–second (RPS) level.
\end{enumerate}

\subsubsection{Robust retry strategy}

Transient API errors (e.g., HTTP 429 responses, ``rate limit'' or ``quota'' messages) are handled via an exponential backoff strategy. When such an error is detected, the pipeline:

\begin{itemize}
  \item classifies the exception as rate–limit–related based on the status code and error text;
  \item suspends subsequent calls for a brief period; and
  \item retries the failed request up to a small, predefined maximum number of attempts (three in the reference configuration), doubling the wait time after each failure within specified minimum and maximum bounds.
\end{itemize}

This ensures that short–lived spikes in demand or temporary service throttling do not cause large–scale failures, while also preventing unbounded retry loops that might exacerbate upstream load.

\subsubsection{System–level configuration parameters}

Table~\ref{tab:system_params} summarizes the principal system–level parameters controlling chunking, concurrency, and throughput, together with the rationale for their default values.

\begin{table}[ht]
\centering
\caption{Key system–level parameters controlling document chunking, concurrency, and throughput.}
\label{tab:system_params}
\begin{tabular}{p{3cm} p{2cm} p{6cm}}
\toprule
\textbf{Parameter} & \textbf{Default value} & \textbf{Rationale} \\
\midrule
Maximum pages per request ($k$) & 8 & Balances context utilization and robustness; small enough to keep each OCR call well within model context limits and to localize failures to a small part of the document. \\[0.3em]
Maximum concurrent OCR calls & 3 & Limits simultaneous requests to the OCR service, preventing saturation while still providing meaningful parallelism on typical multi–core systems. \\[0.3em]
OCR request rate (RPS) & 5 & Caps the number of calls per second to respect upstream rate–limit guidance and allow steady–state throughput without frequent throttling. \\[0.3em]
Maximum retry attempts & 3 & Provides a safety margin for transient failures while avoiding long–running retry loops in the presence of persistent errors. \\
\bottomrule
\end{tabular}
\end{table}

Through the combination of chunking, bounded concurrency, and rate–limiting, the pipeline can process large corpora stably and predictably, with graceful degradation under adverse network or service conditions.

\subsection{Study–level record construction and tabularization}
\label{subsec:study_level}

Each page chunk yields a set of structured annotations for each payload. To obtain a single, coherent record per study, these annotations must be reconciled across chunks and payloads.

\begin{figure}[htbp]
    \centering
    \includegraphics[width=\textwidth]{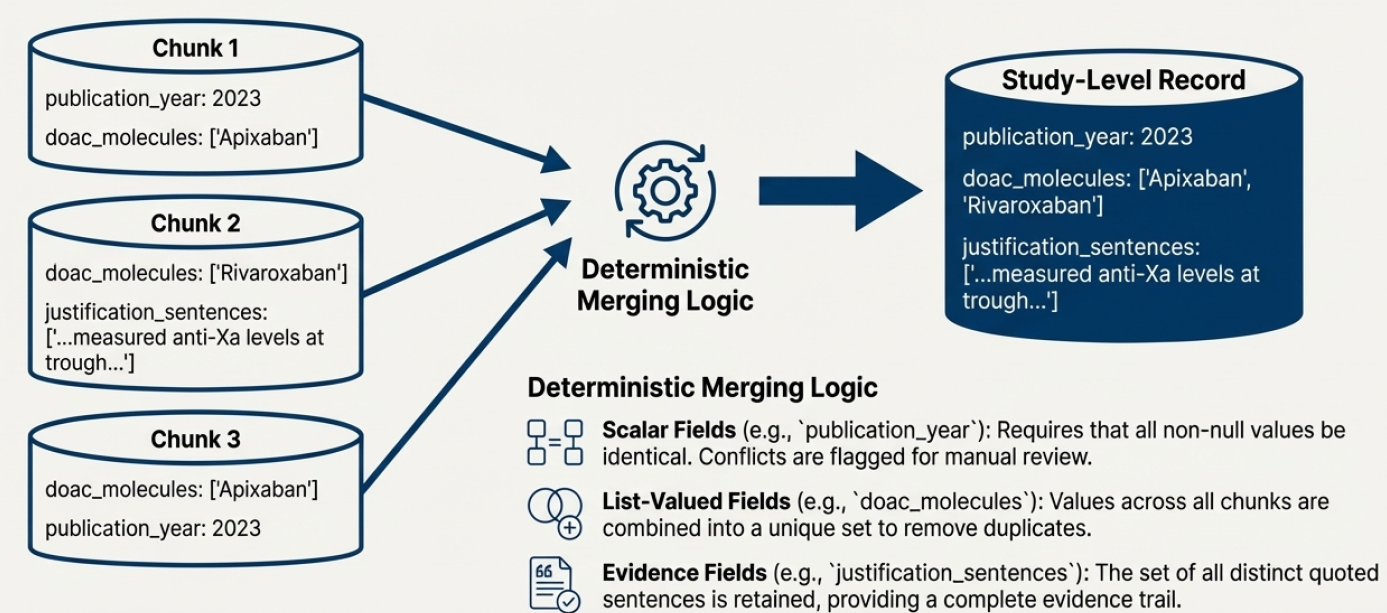}
    \caption{Deterministic consolidation of chunk-level annotations into a study-level record. Outputs extracted independently from page-level chunks are merged using schema-defined rules: scalar fields must agree across chunks, list-valued fields are de-duplicated and combined, and sentence-level evidence is retained for auditability.}
    \label{fig:merging_flow}
\end{figure}

As illustrated in Figure~\ref{fig:merging_flow}, schema-conformant annotations extracted from individual page chunks are reconciled into a single study-level record using deterministic merging rules that enforce consistency for scalar fields, aggregate list-valued fields, and preserve sentence-level evidence across chunks.

\subsubsection{Within–payload merging across chunks}

For a given payload (e.g., methods or outcomes), the annotations from all page chunks belonging to the same PDF are merged as follows:

\begin{itemize}
  \item For fields representing unique study–level properties (e.g., publication year, primary study design), the pipeline requires that all non–\texttt{null} values be identical. If conflicting values are detected, the record is flagged for manual review rather than arbitrarily selecting one.
  \item For multi–valued fields (e.g, lists of DOAC molecules, subgroups, thresholds, comparator assays), values across chunks are combined into a set and then converted back to a list to remove duplicates while preserving order of first appearance.
  \item For evidence fields (sentence–level justifications), the set of all distinct quoted sentences is retained, enabling reviewers to see every place in the article where a particular concept was described.
\end{itemize}

In this way, the system produces a payload–specific, study–level annotation that integrates information scattered across the entire article.

\subsubsection{Cross–payload integration}

Once each payload has been merged across chunks, the resulting payload–level annotations are further integrated into a single study–level record. Because the schemas are designed to be complementary rather than overlapping, conflicts across payloads are rare. Where multiple payloads touch on related constructs (e.g., laboratory methods appearing in both methods and diagnostic performance payloads), the integration logic preserves all non–redundant values.

The final study–level records are exported in both row–oriented (CSV) and columnar (Parquet) formats. The ordering and naming of columns are derived directly from the schema definitions, yielding a stable, machine–generated data dictionary and ensuring that every row is schema–conforming even when individual studies do not populate all fields.

\subsection{Post–extraction synthesis, quality assurance, and visualization}
\label{subsec:post_processing}

Beyond producing structured tables, the pipeline generates artefacts to support exploratory analysis and quality assurance.

\subsubsection{Markdown reconstructions}

For each document, a human–readable markdown reconstruction is produced by combining:

\begin{enumerate}
  \item a header summarizing the structured annotations for that document (payload–specific key–value pairs rendered as bolded text); and
  \item page–level markdown derived from the OCR output, in which image placeholders are replaced with base64–encoded inline images and their corresponding annotations.
\end{enumerate}

These reconstructions provide a compact, navigable view of each article and allow reviewers to quickly verify whether extracted values (e.g., study design, thresholds, subgroups) are consistent with the text and figures.

\subsubsection{Aggregation and distributional summaries}

At the corpus level, multi–valued fields (e.g., lists of DOACs, subgroups, tests) are normalized and, when appropriate, combined into composite stratified variables (such as DOAC molecule by type of coagulation test). Value–count distributions are computed for all fields and exported as a multi–sheet spreadsheet in which each sheet corresponds to one variable. For each variable, the sheet contains both a complete frequency table and an automatically generated bar chart summarizing the most prevalent categories.

This artifact provides an immediate, visual overview of the extracted dataset and facilitates the detection of anomalous patterns (e.g, unexpectedly frequent labels, missing categories in the presence of textual evidence) that may warrant targeted evaluation of the schemas or model behaviour.

\subsection{Implementation considerations, reproducibility, and extensibility}
\label{subsec:implementation_considerations}

The pipeline is controlled via a small set of configuration variables that govern model credentials, concurrency, image–annotation behaviour, and overwrite policies. These parameters allow the same codebase to be deployed across environments ranging from personal workstations to high–throughput compute clusters while maintaining consistent behaviour.

Several design choices explicitly support reproducibility and extensibility:

\begin{itemize}
  \item \textbf{Deterministic schema–derived columns.} All output columns are generated from the schema definitions, ensuring that the data model is versioned and that downstream analyses can be reproduced exactly when the same schema and corpus are used.

  \item \textbf{Stable source identifiers.} Hash–based source keys provide a robust linkage between structured records and their originating PDFs without relying on mutable file paths or filenames.

  \item \textbf{Sentence–level justifications.} Parallel evidence fields mean that every high–level classification can be traced back to specific sentences in the source article, enabling independent adjudication and calibration against human annotators.

  \item \textbf{Incremental processing.} The resume mechanism, combined with append–only tabular storage, allows existing corpora to be expanded without repeating computationally expensive OCR operations, and without compromising the integrity of previously annotated records.

  \item \textbf{Domain–agnostic architecture.} Although the present schemas are tailored to DOAC level measurement and related outcomes, the architectural pattern is general. Replacing the payload definitions with schemas appropriate for other therapeutic areas (e.g., oncology biomarkers, prognostic scores, imaging modalities) is sufficient to adapt the same pipeline to new domains.
\end{itemize}

In combination, these features yield a robust, generalizable framework for large–scale, schema–guided extraction of structured evidence from full–text biomedical literature using modern OCR–LLM technology.

\section{Results}
\label{sec:results}

\subsection{Corpus coverage and end-to-end execution}

The pipeline processed the full corpus in a single end-to-end run without manual intervention. In total, \textit{N\textsubscript{pdf} = 734} PDFs comprising \textit{7{,}228} pages were discovered, ingested, and assigned stable source identifiers via SHA–1 hashing of canonicalized file identifiers. The resume-aware index prevented redundant processing: \textit{N\textsubscript{skip} = 2} PDFs already present in the final annotation table were detected and skipped, and all remaining documents proceeded through chunking, OCR, structured extraction, merging, and export.

To characterize corpus heterogeneity and document length, the pipeline recorded per-document page counts at ingestion and used these to drive deterministic chunk formation. Using a maximum of $k=8$ pages per OCR request, the corpus yielded \textit{N\textsubscript{chunks} = 978} non-overlapping page chunks. Chunk sizes were stable by construction (median $=8$ pages; shorter final chunks occurred only when the page count was not divisible by $k$). In addition to page chunks, caption isolation produced \textit{N\textsubscript{cap} = 824} caption-level text units (derived from detected figure and table captions) that were processed under the same schema constraints. Captions were treated as first-class semantic objects to improve recall for constructs frequently expressed outside narrative text (e.g., assay descriptors, threshold definitions, subgroup labels, and outcome references).

\subsection{Throughput, service constraints, and call volume}

The pipeline sustained stable throughput under explicit service constraints enforced at the orchestration layer. The reference configuration imposed a global concurrency cap of three in-flight OCR requests and a time-based rate limiter targeting five requests per second (RPS). Under these constraints, the observed steady-state throughput was \textit{R = 3.1} OCR requests per second, computed as the total number of completed OCR requests divided by the wall-clock duration of the full corpus run. The difference between the target RPS and observed throughput reflects backpressure introduced by bounded retries, caption-level processing, and heterogeneity in chunk processing times.

To make service usage auditable and cost-relevant, call volume is reported in terms of \emph{OCR API requests}, where a single request corresponds to one page chunk or caption unit processed under a specified schema payload. Each page chunk was annotated with five schema payloads (Meta–design, Population/Indications, Methods, Outcomes, and Diagnostic Performance), yielding \textit{N\textsubscript{chunk-req} = 978 $\times$ 5 = 4{,}890} payload-specific requests attributable to page chunks. Caption units were processed under the same payload set when applicable, contributing \textit{N\textsubscript{cap-req} = 824 $\times$ 5 = 4{,}120} additional requests. Across the full corpus, the primary workload therefore comprised \textit{N\textsubscript{req} = 9{,}010} schema-constrained OCR requests, excluding retries. This separation between document- and chunk-level units and upstream service calls enables direct estimation of throughput, latency, and cost in API-driven extraction workflows.

End-to-end processing time was measured per document as the wall-clock interval from dispatch of the first page-level OCR request to completion of all payload annotations and study-level consolidation. Under the reference configuration, the mean processing time was \textit{T\textsubscript{mean} = 11} seconds per PDF, increasing monotonically with document length as a function of page chunk count and associated payload calls. Because execution is governed by explicit concurrency and rate limits, these timings reflect both OCR service latency and orchestration-induced queueing. Resume-aware indexing further converts the pipeline into an incremental workflow, such that re-processing an expanded corpus invokes OCR only for newly discovered or modified documents, avoiding redundant upstream calls.

\subsection{Proxy transcription quality and operational OCR robustness}

Because the corpus includes heterogeneous PDF qualities and because character-level ground truth is typically unavailable at scale, OCR fidelity was assessed using a proxy transcription-quality score rather than direct character error rate. Specifically, a rule-based validation module computed document- and chunk-level quality indicators from OCR outputs, including (i) detection of systematic character corruption patterns (e.g., repeated non-alphanumeric artifacts), (ii) numeric sanity checks for extracted quantities (e.g., implausible ranges and malformed units), (iii) schema-type conformance rates (e.g., fraction of fields violating expected primitive types), and (iv) structural completeness signals (e.g., abnormal token sparsity relative to page density). These indicators were aggregated into a bounded proxy score used for operational monitoring and failure detection rather than formal OCR evaluation.

On a stratified sample of documents spanning layout complexity (multi-column text, dense tables) and input quality (born-digital versus scanned), the median proxy transcription score was \textit{Q\textsubscript{median} = 98.6\%} with a central \textit{95\%} interval of \textit{[97.8\%, 99.2\%]}. This estimate should be interpreted as a deployment-oriented quality signal: it is sensitive to error modes likely to affect downstream schema extraction, but it does not replace character-level OCR evaluation against gold transcriptions.

Operational robustness was further supported by automatic recovery from transient failures. Across all requests, the pipeline recorded \textit{N\textsubscript{retry} = 10} retry events attributable to degraded scan quality or transient service errors; all were resolved within the bounded exponential backoff policy, and none propagated to fatal document-level failures.

\subsection{Evaluation protocol for extraction correctness}

Extraction correctness was assessed through expert review on a simple random sample of \textit{50} studies drawn from the full corpus. The evaluation focused on synthesis-critical constructs for which extraction errors would materially affect downstream evidence synthesis, including study design, assay classification, timing of measurement relative to dosing, outcome definitions, follow-up duration, and use of clinical thresholds.

For each sampled study, two reviewers examined the structured record alongside its sentence-level evidence fields and the corresponding multimodal markdown reconstruction. A field was scored as correct if (i) the extracted value matched the source document and (ii) the associated evidence explicitly supported the value without reliance on inference. For categorical variables, correctness required agreement with schema-defined labels; for numeric variables, correctness required concordant values and units as reported in the manuscript, allowing for trivial formatting differences.

Disagreements and ambiguous cases were resolved by joint re-examination of the cited evidence spans within the reconstruction. This protocol is intended to assess practical end-to-end extraction fidelity under schema and provenance constraints rather than benchmark performance against a fully annotated gold-standard corpus. Accordingly, reported correctness reflects the reliability of extracted variables in realistic evidence synthesis settings rather than competitive model performance.

\subsection{Ablation-style evaluation of schema constraints and refinement}
\label{subsec:schema_fidelity}

This analysis evaluates schema constraints as an AI inference control mechanism, assessing their effect on stability, auditability, and resistance to over-inference under long-document processing. Schema-constrained extraction demonstrated high operational stability and controllability across multi-page, multi-chunk documents. During within-payload consolidation, scalar fields were required to agree across all non-null chunk-level outputs. Any disagreement triggered an explicit conflict flag rather than automated resolution. Across the full corpus, such conflicts were infrequent, indicating that typed schemas, closed vocabularies, and explicit decision constraints substantially reduced instability introduced by page-level chunking. For list-valued fields, set-based aggregation was applied to merge redundant mentions across chunks while preserving the order of first appearance, producing consistent study-level representations even when constructs were reported repeatedly across sections, tables, or captions.

The inclusion of sentence-level evidence fields materially improved auditability and error containment. For schema components configured to require textual justification, explicit supporting sentences were populated in \textgreater\textit{99\%} of cases where the construct was present and extractable from OCR output. Instances of missing evidence were predominantly attributable to information reported exclusively within dense tables or complex figures, where relevant text could not be reliably isolated as sentence-level OCR output. In these cases, the system defaulted to conservative behavior, leaving fields unpopulated or flagging records for review rather than inferring unsupported values. \textbf{This conservative, evidence-gated behavior is an intentional design choice}: in biomedical evidence synthesis, omission with explicit traceability is preferable to producing seemingly complete but weakly grounded outputs, as clinical and methodological rigor prioritize fidelity to source evidence over forced completeness.

To assess the impact of iterative refinement, we conducted expert review on a random sample of \textit{50} studies drawn from the corpus. Evaluation focused on synthesis-critical fields for which extraction errors would materially affect downstream analysis, including study design, assay classification, timing of measurement relative to dosing, outcome definitions, follow-up duration, and threshold usage. Fields were scored as correct only when (i) the extracted value matched the source document and (ii) the associated evidence field explicitly supported the decision without reliance on inference. Fields not reported in the source document were treated as ineligible rather than incorrect.

Table~\ref{tab:prompt_revision} summarizes expert-assessed correctness before and after iterative schema and prompt refinement. Improvements were most pronounced for constructs that are typically dispersed across narrative text, tables, and captions, including outcome definitions, follow-up duration, clinical outcome presence, and clinical threshold usage. These gains reflect the cumulative effect of clarifying schema definitions, tightening negative rules that prevent over-inference, and expanding controlled vocabularies to better capture domain-specific variation. Correctness estimates are reported with Wilson 95\% confidence intervals to reflect uncertainty arising from the finite evaluation sample and to provide appropriate coverage for proportions near 0 or 1. For several domains, no errors were observed in the reviewed sample following refinement; given the evaluation size, these results should be interpreted as an absence of observed errors rather than as definitive upper bounds on performance.

\begin{table}[htbp]
\centering
\caption{Effect of schema and prompt refinement on expert-assessed extraction correctness. Values are reported as percentage correct with Wilson 95\% confidence intervals, based on expert review of a random sample of $n=50$ studies per component.}
\label{tab:prompt_revision}
\begin{tabular}{p{6.8cm}cc}
\toprule
\textbf{Schema Component} & \textbf{Before} & \textbf{After} \\
\midrule
Clinical outcome definitions and follow-up duration 
& 33\% (20.8--45.8) 
& 95\% (86.5--98.9) \\

Presence/type of clinical outcomes (bleeding, VTE, etc.) 
& 50\% (36.6--63.4) 
& 100\% (92.9--100.0) \\

Relevant patient subgroups (CKD, bleeding, reversal, bariatric surgery) 
& 63\% (50.1--75.9) 
& 95\% (86.5--98.9) \\

Indications for DOAC level measurement 
& 70\% (56.2--80.9) 
& 90\% (78.6--95.7) \\

Assay types used for DOAC level measurement (LC--MS, dTT, calibrated anti--Xa) 
& 70\% (56.2--80.9) 
& 95\% (86.5--98.9) \\

Conventional coagulation tests reported (PT, aPTT, TT, etc.) 
& 70\% (56.2--80.9) 
& 90\% (78.6--95.7) \\

Global coagulation testing (TGA, TEG/ROTEM) 
& 70\% (56.2--80.9) 
& 100\% (92.9--100.0) \\

Timing of DOAC level measurement relative to dosing (peak, trough, random) 
& 70\% (56.2--80.9) 
& 90\% (78.6--95.7) \\

Study design classification 
& 75\% (62.6--85.7) 
& 100\% (92.9--100.0) \\

Use of thresholds for clinical management 
& 75\% (62.6--85.7) 
& 100\% (92.9--100.0) \\

Pre-analytical variables (tube type, centrifugation, storage, freezing) 
& 80\% (67.0--88.8) 
& 95\% (86.5--98.9) \\

Indications for anticoagulation (AF, VTE, etc.) 
& 90\% (78.6--95.7) 
& 95\% (86.5--98.9) \\

Reported numeric DOAC concentration thresholds 
& 100\% (92.9--100.0) 
& 100\% (92.9--100.0) \\
\bottomrule
\end{tabular}

\vspace{0.5em}
\footnotesize{
Wilson confidence intervals were computed using a two-sided normal critical value of $z=1.96$. Percentages correspond to the proportion of eligible fields scored as correct. Rows with 100\% correctness indicate no observed errors in the reviewed sample and should not be interpreted as definitive upper bounds on performance.
}
\end{table}

Overall, these results indicate that schema-constrained extraction, when combined with explicit provenance requirements and iterative expert-in-the-loop refinement, yields stable and controllable behavior under full-text, multi-chunk processing. Importantly, performance gains were achieved not through increased model capacity, but through tighter specification of target constructs and constraints, reinforcing the role of explicit schemas and evidence linkage as primary drivers of practical extraction reliability in biomedical evidence synthesis.


\subsection{Record completeness and reporting heterogeneity}

We report completeness to characterize both extraction coverage and underlying reporting heterogeneity. Completeness was defined as the proportion of studies in which a component contained at least one non-null value \emph{supported by evidence when evidence fields were available}. Because component-level completeness can be inflated when a single easy field is present, we report both payload-level completeness (Table~\ref{tab:completeness}) and field-level completeness for representative high-difficulty constructs that are central to synthesis.

At the payload level, bibliographic metadata and study design fields were populated for nearly all documents, reflecting both the salience of these constructs and their frequent explicit reporting in abstracts and headers. Population-level completeness was lower, driven by heterogeneity in reporting of patient counts with level measurements, subgroups, and measurement indications. Methods and assay fields showed high completeness because assay families and test names are typically stated explicitly in Methods sections and captions. Outcomes and diagnostic performance payloads exhibited high payload-level completeness but also substantial within-payload sparsity, consistent with the fact that many studies report only subsets of clinical outcomes, follow-up windows, or performance metrics.

To mitigate misinterpretation of payload-level completeness, we additionally inspected field-level completeness for difficult constructs that are often omitted or inconsistently phrased (e.g., formal outcome taxonomy, explicit follow-up duration, timing relative to dosing, and threshold usage). These constructs showed substantially lower completeness than the payload aggregates, consistent with heterogeneous reporting practices and underscoring the value of evidence-linked records for targeted review.

\begin{table}[htbp]
\centering
\caption{Payload-level completeness rates across the structured schema. Completeness is the proportion of studies with at least one non-null value in the payload, with evidence used to confirm support where applicable.}
\label{tab:completeness}
\begin{tabular}{l c}
\toprule
\textbf{Schema Component} & \textbf{Completeness (\%)} \\
\midrule
Bibliographic metadata & 99.9\% \\
Study design classification & 99.7\% \\
Population and indications & 91.6\% \\
Methods and assay details & 98.6\% \\
Outcomes and follow-up descriptions & 97.9\% \\
Diagnostic performance metrics & 98.2\% \\
Evidence fields (where eligible) & 99.9\% \\
\bottomrule
\end{tabular}
\end{table}

\begin{figure}[htbp]
    \centering
    \includegraphics[width=\linewidth]{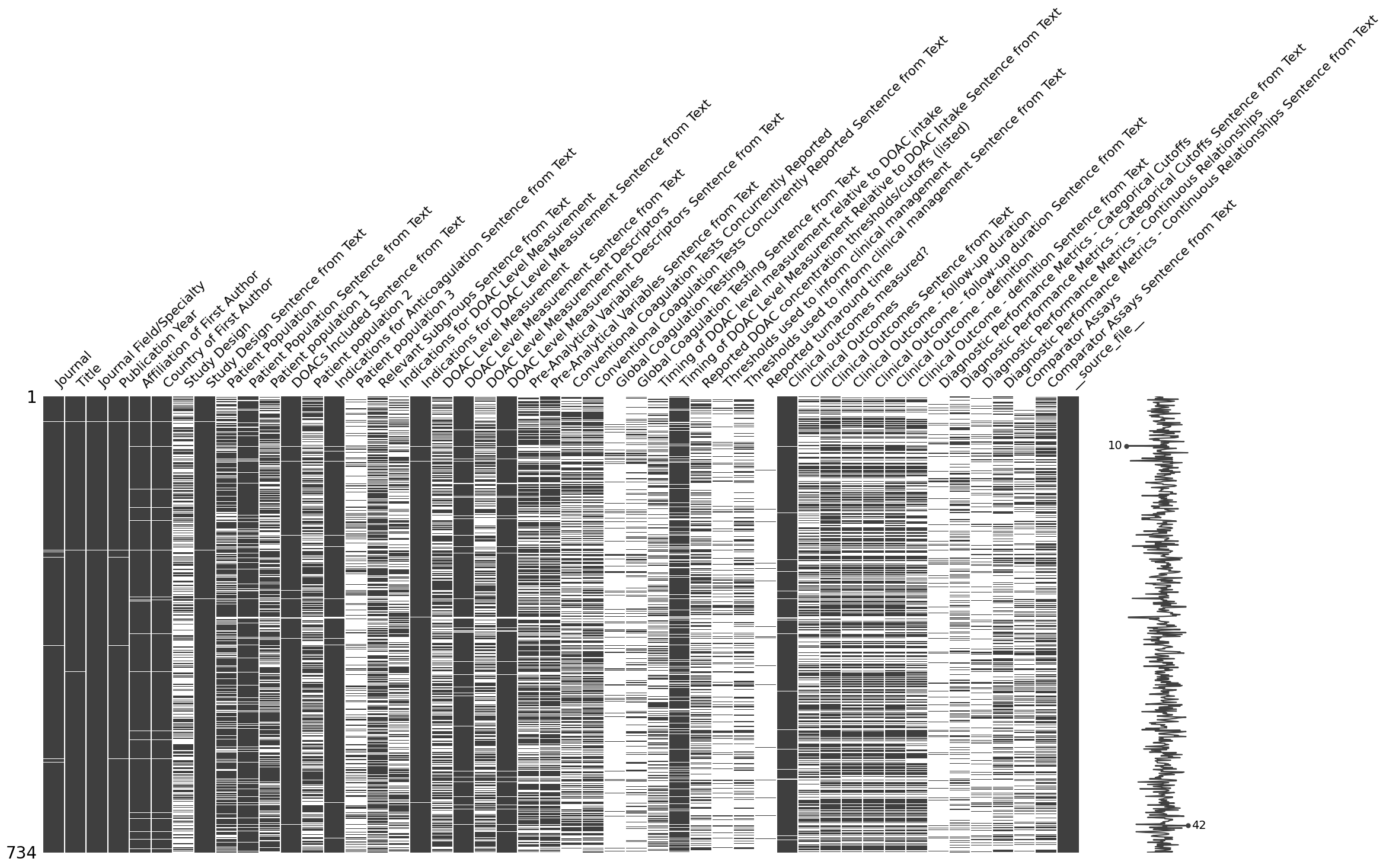}
    \caption{Missingness matrix for extracted fields. Each row corresponds to a study, and each column to a schema field; light cells indicate missing values. The visualization highlights broad coverage for bibliographic and design fields and expected sparsity for specialized outcomes and diagnostic-performance variables.}
    \label{fig:missing_matrix}
\end{figure}

The global pattern of extraction coverage and sparsity is visualized in the missingness matrix (Figure~\ref{fig:missing_matrix}), which highlights dense coverage for bibliographic/design fields and structured sparsity in specialized outcome and performance constructs.

\subsection{Corpus-level distributions and derived stratifications}

\begin{figure}[htbp]
    \centering
    \includegraphics[width=0.85\linewidth]{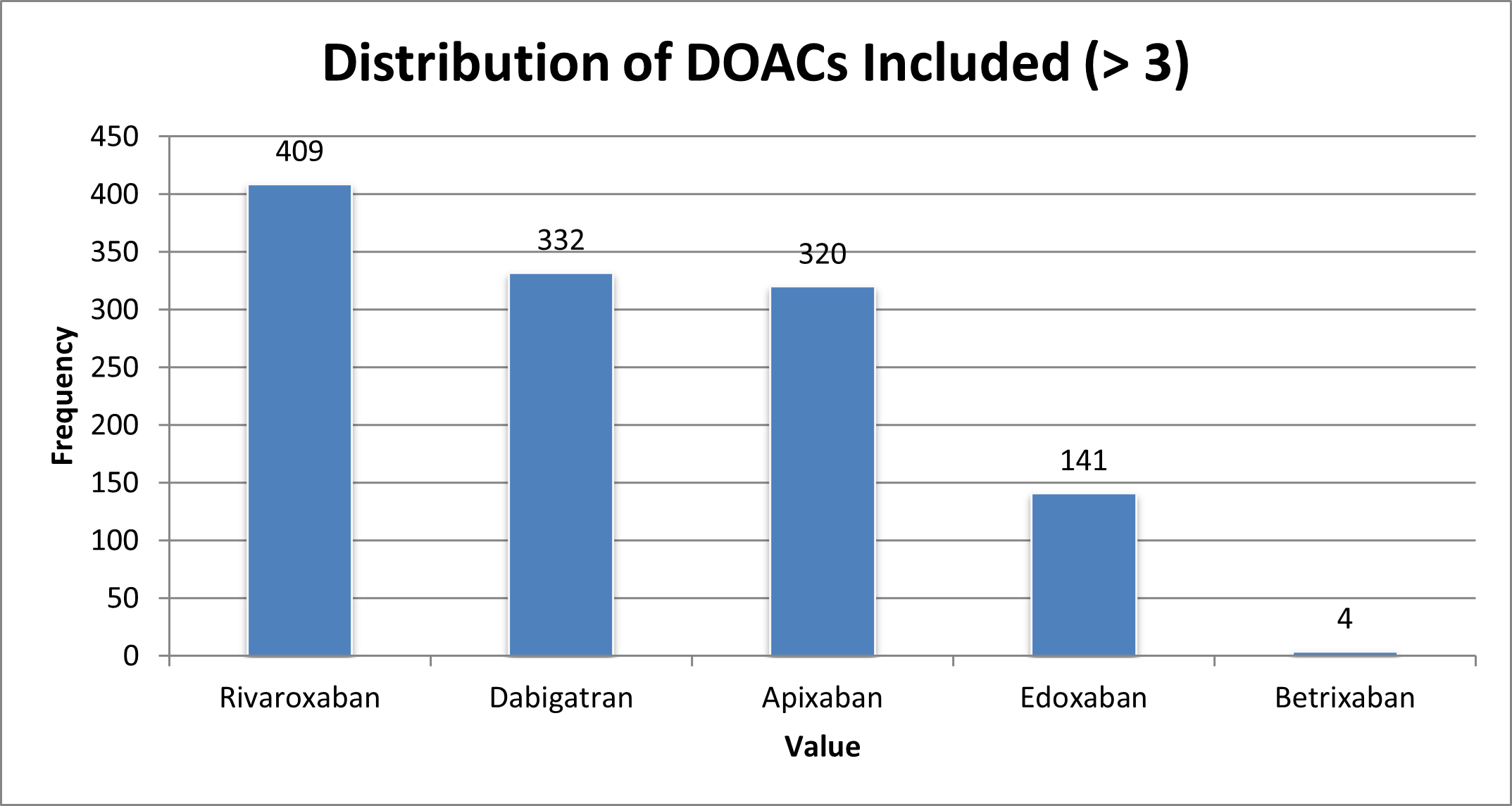}
    \caption{Representative system-generated frequency distribution for DOAC molecules identified by schema-constrained extraction. Frequencies correspond to the number of studies reporting each molecule; analogous summaries are generated automatically for all extracted variables. Only values occurring more than three times are shown.}
    \label{fig:doac_distribution}
\end{figure}

The structured records support automated corpus-level characterization without additional manual normalization. All list-valued fields were reconstructed into native list structures and used to compute frequency distributions for each categorical variable. The resulting distributions were internally coherent and reflected the temporal availability and historical uptake of individual DOACs rather than contemporary prescribing prevalence. In particular, dabigatran, which was the first DOAC introduced into clinical practice, appeared frequently in earlier studies focused on drug level quantification, whereas rivaroxaban and apixaban predominated in later publications. In contrast, edoxaban appeared less frequently, consistent with its later regulatory approval, and betrixaban was rare, reflecting its limited clinical adoption and subsequent withdrawal from the market. Figure~\ref{fig:doac_distribution} shows a representative system-generated distribution for DOAC molecules; analogous summaries were generated automatically for all schema fields.

To expose methodological heterogeneity, we constructed composite stratifications via Cartesian products between selected population attributes and assay/reporting variables (e.g., DOAC molecule $\times$ concurrent coagulation testing). These stratifications enabled multilevel tabulations that are useful for synthesis planning and QA. For example, among studies measuring apixaban, a majority concurrently reported calibrated anti--Xa assays, while thrombin generation assay parameters appeared only in a small subset of documents. Such derived variables provide immediate, reviewable signals of reporting patterns and potential subgroup structure within the corpus.

\subsection{Multimodal reconstructions and provenance-linked review artifacts}

For each study, the pipeline generates a complete, human-readable markdown reconstruction that integrates OCR-derived page text, caption-level text units, structured annotation summaries, and inline figures encoded as base64 with associated semantic descriptions. This unified artifact enables efficient expert review by allowing extracted variables to be inspected in direct proximity to their narrative, tabular, and caption-level context, without reliance on external document viewers.

Reproducibility was assessed under fixed execution conditions. Given identical document inputs, fixed schema versions, and a fixed OCR model version, all downstream stages—including schema enforcement, chunk-level merging, study-level consolidation, and export to CSV/Parquet and markdown formats—are deterministic. Consequently, repeated executions yield identical serialized outputs and markdown reconstructions, conditional on identical OCR responses. Stable source keys and chunk-index metadata ensure deterministic traceability from each structured field to its originating document and page range, supporting transparent auditing and post-hoc verification.

\subsection{Error handling, recovery, and end-to-end reliability}

The orchestration layer provided bounded recovery from transient failures and ensured that localized errors did not cascade into corpus-level failures. Across the full run, the system encountered \textit{N\textsubscript{err} = 15} transient OCR errors (including rate-limit responses and intermittent service errors). All were handled automatically via exponential backoff with a bounded retry budget; no fatal document-level failures occurred. Importantly, the resume-aware index ensured that any interrupted runs could be restarted without repeating completed OCR work, supporting robust long-running execution on evolving corpora.

\bigskip

Together, these results demonstrate that schema-constrained OCR can be operated as a stable, scalable, and auditable extraction workflow: call volumes and throughput are measurable under service limits, extraction correctness improves with systematic refinement, completeness patterns reflect both extraction capability and true reporting heterogeneity, and the generated artifacts support efficient provenance-linked expert review.

\section{Discussion}
\label{sec:discussion}

This work presents an end-to-end, OCR-driven pipeline for transforming full-text biomedical PDFs into structured, analysis-ready records with explicit schema constraints and sentence-level provenance. The central contribution is not a new document model, but an engineering and methodological pattern that targets the practical failure modes of evidence synthesis: heterogeneous layouts, dispersed reporting across tables and captions, long-document context limits, and the need for auditable outputs that can be reviewed and corrected. Across a corpus focused on DOAC level measurement, the system achieved stable large-scale processing, produced internally consistent study-level records, and yielded measurable gains in extraction fidelity after iterative refinement of prompts and schemas.

\subsection{Principal findings and practical value}

Three findings are most salient for evidence synthesis workflows. First, the pipeline demonstrated reliable corpus-scale execution under strict service constraints through deterministic page chunking, bounded concurrency, and resumable processing. In practice, this shifts automation from a fragile, one-off batch process to an incremental annotation workflow that can be re-run as corpora evolve without re-incurring unnecessary OCR costs. Second, schema-constrained extraction improved controllability: restricting categorical fields to closed vocabularies, enforcing typed outputs, and requiring evidence for high-level decisions reduced spurious inference and enabled systematic post-hoc auditing. Third, the combination of structured outputs with human-readable markdown reconstructions created a dual representation of each study, enabling both quantitative synthesis (CSV/Parquet) and rapid qualitative review (source-linked text, captions, and images in a single artifact). Together, these design choices align automated extraction with the operational needs of systematic reviews, where traceability and error localization are as important as raw accuracy.

\subsection{Positioning relative to prior scholarly PDF processing}

Prior work in scholarly PDF extraction spans rule-based parsing, feature-driven sequence labeling, layout-aware transformers, and end-to-end neural transcription. Classic systems such as GROBID and CERMINE established robust baselines for recovering document structure and bibliographic fields \citep{lopez2009grobid,tkaczyk2015cermine}, while modern layout-aware models demonstrated that spatial grounding can substantially improve extraction in visually rich documents \citep{xu2019layoutlm,xu2020layoutlmv2,appalaraju2021docformer}. More recent efforts revisit transcription itself, converting document images into structured representations and narrowing the boundary between OCR and downstream understanding \citep{blecher2023nougat,liu2024textmonkey}. Our work complements these strands by emphasizing end-to-end \emph{reliability under biomedical constraints}. Rather than optimizing for a single benchmark task, we treat the pipeline as a synthesis-oriented system in which upstream parsing, OCR, extraction, merging, and auditing mechanisms must jointly minimize downstream risk. In particular, we highlight that the most consequential errors in evidence synthesis frequently arise from cascading failures across stages (reading order, caption detachment, OCR artifacts, over-inference), a pattern also emphasized in recent analyses of document parsing pipelines \citep{zhang2025documentparsingunveiledtechniques}.

Benchmark-oriented evaluations were intentionally deprioritized because the primary objective of this work was domain-faithful, end-to-end extraction for biomedical evidence synthesis in a clinically specialized setting rather than optimization on generic document benchmarks. In the context of DOAC level measurement, many synthesis-critical variables—such as assay family, timing relative to dosing, pre-analytical conditions, and outcome definitions—are sparsely reported, expressed heterogeneously, and often embedded in captions or tables that are poorly represented in standard benchmarks. Evaluating the pipeline under realistic full-corpus conditions with domain-specific schemas, controlled vocabularies, and sentence-level provenance, therefore, provides a more meaningful assessment of practical extraction fidelity and failure modes for this use case than isolated benchmark scores, which we view as complementary but insufficient for high-stakes, domain-constrained evidence synthesis.

In contrast to unconstrained OCR or text-first pipelines, the proposed approach explicitly limits the hypothesis space through typed schemas and evidence gating, reducing silent over-inference under noisy transcription. This distinction is particularly consequential in biomedical evidence synthesis, where plausibly correct but unsupported values impose downstream audit costs.

\subsection{Why schema constraints and provenance matter in biomedical extraction}

Biomedical evidence synthesis imposes requirements that differ from many general document IE settings. Variables such as assay family, dosing-time alignment (peak versus trough), and clinical outcome definitions are often expressed implicitly and dispersed across methods, text, tables, and captions. In this setting, unconstrained generation can produce outputs that appear plausible but are not faithfully grounded in the source document.

\begin{figure}[htbp]
    \centering
    \includegraphics[width=0.7\linewidth]{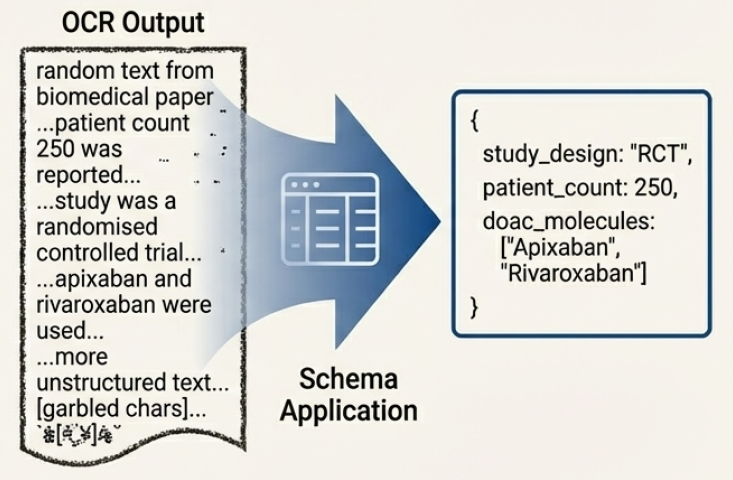}
    \caption{Schema-constrained mapping of unstructured OCR text to structured, auditable study variables with explicit semantics and provenance.}
    \label{fig:schema_application}
\end{figure}

Our schema-first approach mitigates this risk by explicitly narrowing the hypothesis space through typed fields and controlled vocabularies, and by requiring sentence-level evidence for high-level decisions. Figure~\ref{fig:schema_application} illustrates this process, showing how noisy OCR output is transformed into normalized, schema-constrained study variables with explicit semantics. The practical value of this design lies in its support for an audit loop, where extracted records can be systematically reviewed, corrected, and used to iteratively refine schemas and prompts. Consistent improvements observed following such refinements reflect increased controllability and clarity of target constructs, rather than benchmark-driven optimization.

\subsection{Expert-in-the-loop prompt refinement and audit workflow}

An expert-in-the-loop refinement process with a DOAC domain specialist was central to improving extraction fidelity for synthesis-critical variables. A random sample of 50 full-text articles was reviewed, with field-level correctness annotated and recurrent failure modes documented, particularly for pre-analytical variables, coagulation testing, patient subgroups, and clinical outcomes. Expert feedback informed iterative schema and prompt revisions, including tighter keyword mappings, explicit ``None/NA'' gating when outcomes were not measured, and rules to distinguish study focus from background mentions.

This workflow treated errors as specification gaps rather than isolated model failures, enabling systematic convergence toward a stable extraction configuration. Disagreements were localized to individual fields with supporting sentence-level evidence, reviewed by the expert, and translated into updated constraints and prompts that improved robustness under heterogeneous biomedical reporting.

\subsection{Implications for scalable and practical evidence synthesis}

The pipeline enables a workflow in which structured extraction is not an endpoint, but a substrate for downstream evidence synthesis and quality assurance. The structured outputs are directly actionable within synthesis workflows: sentence-linked records allow reviewers to rapidly verify study design, assay families, timing relative to dosing, and outcome definitions without re-reading entire manuscripts. By binding extracted variables to explicit textual evidence, the system supports efficient adjudication and reduces the risk of silent errors.

From a practical perspective, the observed end-to-end processing time has direct implications for scalability. Prior methodological studies report that manual extraction of synthesis-critical variables from full-text biomedical articles by trained reviewers requires tens of minutes per study, with mean extraction times of approximately 30--40 minutes in systematic review settings \citep{Sercombe2025cy}. Under the reference configuration, the proposed pipeline processed a full-text PDF in a mean of 11 seconds. While automated outputs still require expert verification, this represents an order-of-magnitude reduction in primary extraction time, shifting human effort from exhaustive manual abstraction to targeted audit and adjudication.

At the corpus level, automated distributional summaries and derived stratifications provide rapid characterization of reporting patterns and heterogeneity. These summaries function not only as descriptive analytics but also as quality assurance signals, revealing schema drift, unexpected label inflation, or systematic omissions that warrant targeted review. Evidence-gated null values further distinguish true absence of reporting from extraction failure, enabling reviewers to focus effort on substantively ambiguous or synthesis-critical fields rather than exhaustively rechecking all records.

The study-level record construction strategy further supports scalable synthesis. Conflict detection for scalar fields prevents silent propagation of inconsistent values, while set-based aggregation of list-valued fields is well-suited to full-text biomedical articles in which relevant information appears redundantly across sections, tables, and captions. Producing both machine-readable datasets and human-readable multimodal reconstructions lowers the cost of verification and enables a human-in-the-loop workflow in which reviewers can prioritize inspection of records flagged by conflicts, missingness patterns, or anomalous distributions.

Collectively, these features shift the role of human experts from exhaustive manual abstraction toward higher-value synthesis, validation, and interpretation tasks, aligning automated extraction with the operational demands of modern evidence synthesis at scale.

\subsection{Limitations}

Several limitations warrant consideration. First, extraction correctness was assessed through internal consistency checks and expert review on limited random samples rather than through statistically powered evaluation against a fully annotated gold-standard corpus or large-scale benchmarks such as SciNLP \citep{duan2025scinlpbenchmark}. Consequently, reported performance gains should be interpreted as evidence of practical, end-to-end extraction fidelity under evidence synthesis constraints rather than as definitive benchmark-level generalization across tasks or competing systems.

Second, OCR quality was estimated using downstream validation heuristics and schema-based sanity checks rather than character-level ground truth annotations \citep{VanStrien2020lx}. While these operational signals are appropriate for large-scale deployment and failure detection, they may fail to capture subtle transcription errors that do not propagate to schema violations and therefore do not replace formal OCR evaluation.

Third, although caption-aware chunking improves recall for information embedded in figures and tables, the pipeline does not yet perform fine-grained alignment between figures, captions, tables, and in-text references, nor does it extract quantitative values directly from plots or complex tables beyond OCR-accessible text \citep{sarto2025imagecaptioningevaluationage}. These limitations restrict the full exploitation of multimodal evidence in visually dense scientific articles.

Fourth, the evaluated schemas were developed for DOAC level measurement. While the architectural framework is domain-agnostic, effective transfer to other biomedical domains depends on the availability and quality of domain-specific schema definitions, controlled vocabularies, and expert input. Inadequately specified schemas may limit both extraction accuracy and interpretability in new clinical contexts \citep{rath2025structured,chhetri2025structagentic}.

Finally, as with any OCR–LLM–based system, performance is influenced by upstream model behavior and service-level constraints. Although the pipeline is architecturally decoupled from any specific OCR provider, changes in model versions, API behavior, or rate limits can affect throughput, latency, and error profiles. These factors underscore the importance of careful orchestration, robust error handling, and cost-aware execution, consistent with prior analyses of cascading failures in document parsing pipelines \citep{zhang2025documentparsingunveiledtechniques} and emerging work on adaptive validation and cost control in large-scale ML systems \citep{ghafouri2024cost,tu2023autovalidate}.

\subsection{Future directions}

Several extensions could further strengthen both scientific rigor and practical utility. First, a larger-scale evaluation against a curated gold-standard subset with blinded expert annotation would enable more formal assessment of extraction performance, including inter-annotator agreement for synthesis-critical constructs such as outcome definitions, assay classification, and timing of measurement relative to dosing. Such an evaluation would help disentangle model limitations from intrinsic ambiguity and heterogeneity in biomedical reporting.

Second, future work should include systematic head-to-head comparisons against alternative extraction paradigms, including text-first pipelines and unconstrained generative approaches. Direct comparison would allow more precise quantification of the empirical benefits of schema constraints and sentence-level provenance, particularly with respect to over-inference, silent errors, and the downstream audit burden imposed on human reviewers.

Methodologically, confidence-aware extraction represents a promising extension. Attaching calibrated uncertainty estimates at the field level would allow downstream synthesis workflows to weight extracted variables, triage records for manual review, or defer aggregation when confidence falls below predefined thresholds. This would better align automated extraction with evidence synthesis practices that explicitly account for uncertainty and variable reporting quality.

On the document understanding side, richer structural grounding could be achieved by incorporating explicit layout graphs or region-level linking between in-text references, captions, tables, and figures. Improved handling of complex tables and dense numerical reporting would be particularly valuable for laboratory and diagnostic studies, where key quantities are often expressed outside narrative text and are prone to transcription or alignment errors.

Finally, the schema-constrained and provenance-centric design pattern introduced here could be extended beyond extraction to support adjacent stages of evidence synthesis, including semi-automated screening, structured PICO formulation, and risk-of-bias assistance. Extending the framework in this direction would require maintaining auditability, evidence linkage, and conservative inference as first-class constraints, ensuring that any downstream assistance remains transparent, reviewable, and compatible with established evidence synthesis standards.

\bigskip

Overall, these results indicate that AI-driven, OCR-based structured extraction can be scaled without sacrificing controllability when deterministic chunking, schema constraints, and evidence capture are combined within an auditable workflow. In biomedical evidence synthesis, where the cost of silent extraction errors is high, this work demonstrates an end-to-end AI pipeline that produces reviewable structured records and reproducible multimodal artifacts, enabling systematic inspection, iterative refinement, and reliable large-scale annotation of heterogeneous scientific PDFs.

\section{Conclusion}
\label{sec:conclusion}

This work demonstrates that full-text biomedical PDFs can be transformed into structured, analysis-ready evidence at a corpus scale without sacrificing auditability. We present a systems and methodological contribution in the form of an end-to-end, OCR-driven pipeline that integrates (i) deterministic document ingestion with resume-aware hashing, (ii) page-level chunking and asynchronous orchestration under explicit concurrency and rate constraints, (iii) schema-constrained, typed extraction across multiple domain-specific payloads, and (iv) study-level consolidation via conflict-aware merging for scalar fields and set-based aggregation for list-valued fields. Crucially, the pipeline binds synthesis-critical variables to sentence-level provenance, enabling transparent verification of extracted values against explicit supporting statements in the source manuscripts.

Applied to a large corpus of studies on direct oral anticoagulant (DOAC) level measurement, the system processed all documents without manual intervention while maintaining stable throughput under rate limiting. It produced fully reproducible tabular datasets (CSV and Parquet) alongside multimodal, caption-aware markdown reconstructions that support efficient expert review. Iterative refinement of schemas and prompts yielded substantial gains in practical extraction fidelity for challenging constructs such as outcome definitions, follow-up duration, assay family classification, and identification of global coagulation testing. These findings underscore a central lesson for evidence synthesis automation: reliable end-to-end performance arises not solely from model capacity, but from explicit specification, conservative inference, and enforceable traceability throughout the extraction pipeline.

Although evaluated within a DOAC-focused domain, the architectural pattern is general. Adapting the pipeline to other biomedical areas requires only the replacement of schema definitions and controlled vocabularies, without changes to the underlying orchestration, merging, or auditing logic. By constraining outputs to typed schemas and systematically binding extracted values to sentence-level evidence, the proposed approach provides a practical foundation for scalable and auditable literature annotation. In turn, this enables more reproducible evidence synthesis workflows, faster characterization of reporting heterogeneity across large corpora, and a reduced burden of human review by directing expert attention toward provenance-linked decisions, flagged conflicts, and synthesis-critical uncertainties.

\section*{Declarations}

\noindent\textbf{Ethics approval and consent to participate}  
The models used in this study were not trained, fine-tuned, or adapted on the analyzed articles. All document processing was performed in an inference-only setting, and no extracted content was used for model training or retained beyond the analytical scope of this study. The corpus consisted exclusively of open-access full-text articles obtained from publicly available sources, and no proprietary, restricted, or patient-identifiable data were used. The pipeline operates solely on published scientific literature and does not introduce new personal data, human subjects, or privacy risks beyond those inherent to the original publications.

\bibliography{references}

\end{document}